\documentclass{article}

\usepackage[preprint]{neurips_2022}

\usepackage[utf8]{inputenc}
\usepackage{amsmath,amsthm,amssymb}
\usepackage{fullpage}
\usepackage{float}
\usepackage{array, makecell} %
\usepackage{xcolor,colortbl}
\usepackage{framed,color}
\usepackage{enumerate}
\usepackage{cleveref}
\usepackage{comment}
\usepackage{subcaption}

\usepackage{algorithm}
\usepackage{algorithmic}

\usepackage{graphicx}

\usepackage{natbib}

\setcitestyle{numbers}

\newcommand{\etalchar}[1]{$^{#1}$}

\newtheorem{theorem}{Theorem}

\newtheorem{definition}{Definition}

\newtheorem{observation}{Observation}

\newtheorem{fact}{Fact}

\newtheorem{remark}{Remark}
\usepackage{amsmath}

\newcommand{\bui}{\boldsymbol{u_i}}

\newcommand{\mD}{\mathcal{D}}

\DeclareMathOperator*{\argmax}{arg\max}

\newcommand{\bu}{\boldsymbol{u}}
\newcommand{\bv}{\boldsymbol{v}}

\newcommand{\mat}[1]{\boldsymbol{#1}}
\newcommand{\hA}{\hat{A}}

\newcommand{\cred}[1]{\textcolor{red}{\bf\small #1}}

\definecolor{coral}{RGB}{255,187,162}
\definecolor{teall}{RGB}{0,212,212}

\newcolumntype{g}{>{\columncolor{coral}}c}
\newcolumntype{r}{>{\columncolor{teall}}c}

\renewcommand{\deg}{\mathrm{deg}}

\title{Confident Clustering via PCA Compression Ratio and Its
Application to Single-cell RNA-seq Analysis}

\author{
Yingcong Li
\\
Division of Biological Sciences, UC San Diego \\ 
\& La Jolla Institute for Immunology \\
\texttt{yingcongli@lji.org}
\And
Chandra Sekhar Mukherjee 
\thanks{Research supported by NSF CAREER award 2141536.}
\\
Department of Computer Science\\
University of Southern California\\
\texttt{chandrasekhar.mukherjee@usc.edu}
\AND
Jiapeng Zhang
\thanks{Research supported by NSF CAREER award 2141536.}
\\
Department of Computer Science\\
University of Southern California\\
\texttt{jiapengz@usc.edu}
}

\date{\today}

\begin{document}

\maketitle

\begin{abstract}
Unsupervised clustering algorithms for vectors has been widely used in the area of machine learning. Many applications, including the biological data we studied in this paper, contain some boundary datapoints which show combination properties of two underlying clusters and could lower the performance of the traditional clustering algorithms. We develop a confident clustering method aiming to diminish the influence of these datapoints and improve the clustering results. Concretely, for a list of datapoints, we give two clustering results. The first-round clustering attempts to classify only pure vectors with high confidence. Based on it, we classify more vectors with less confidence in the second round. We validate our algorithm on single-cell RNA-seq data, which is a powerful and widely used tool in biology area. Our confident clustering shows a high accuracy on our tested datasets. In addition, unlike traditional clustering methods in single-cell analysis, the confident clustering shows high stability under different choices of parameters. 
\end{abstract}

\section{Introduction}

In this paper, we study unsupervised clustering for high dimensional vectors, with an experimental focus on single-cell RNA-seq data.
We design an algorithm with a concept of ``confidence'' and compare its accuracy and robustness with the state of the art.
We begin by describing the framework of the problem, followed by the data that we work with. Then against that background, we describe the motivations of confident clustering and characteristics of our algorithm. 

\paragraph{Vector clustering problems}
The vector clustering problem is one of the most fundamental problem in machine learning, with a wide range
of applications. In this problem, given a set of vectors we hope 
to cluster them according to their ``similarities''. Famous vector clustering algorithms include KMeans, $k$-NN and DBSCAN.

Concretely, we are given a set $S$ consisting of $n$ many $d$ dimensional vectors, with hidden ground truth clustering $V=\{V_1, \ldots ,V_k\}$.
That is, $S= \cup_{i=1}^k V_i$ and $V_i \cap V_j =\emptyset$ for all $i \ne j$. 
We also use a $d \times n$ matrix $\hA$ to represent the dataset, 
where each row is a feature, and each column is a vector. Given $\hA$, 
we want to obtain a clustering that is close to the hidden clustering $V$.

In this paper we design a confident clustering algorithm that is motivated from single-cell sequencing data, which is one of the recent breakthroughs in biology. Even though our motivation comes from biological data, our algorithm is not restricted to biological domain, as we do not use any biological knowledge in our algorithm design.
(we are interested in datasets that exhibit PCA's compressibility effect~\cite{mukherjee2022compressibility}). 
We also believe this motivation can also be applied in other domains of clustering problems.
Now we provide some background for this datatype.

\subsection{Single-cell analysis and its importance in biology}

Though almost all the cells of a multi-cellular organism share the same set of genome, which contains tens of thousands of genes, they show high diversity in their structure, functions, behaviors,etc. This heterogeneity of cells is achieved by strictly controlled gene expression and regulation. For example, human genome contains more than 60,000 genes, but each single cell only expresses only a few thousands. Even well defined cell populations could still show heterogeneity and have subpopulations. Understanding the gene expression patterns of single cells makes it possible to reveal unknown mechanisms more precisely. However, conventional techniques like fluorescence-activated cell sorting (FACS) or fluorescent microscopy only allow quantification of a few genes at single cell level, in which massive features of gene expression are missed. With the development of single-cell RNA sequencing technique, expression of more than 1000 genes of a single cell can be measured, which makes it a powerful tool to dissect complicated biological processes.

\begin{comment}
\textcolor{red}{However, conventional techniques like fluorescence-activated cell sorting (FACS) or fluorescent microscopy only allow quantification of a few genes, in which massive features of gene expression are missed. With the development of single-cell RNA sequencing technique, expression of more than 1000 genes can be measured at single cell level, which makes it a powerful tool to dissect complicated biological processes.}
\end{comment}

\paragraph{Single-cell data as vector clustering}
In this setup, we have a set of $n$ cells, $S$, for which we have the gene-expression of the same $d$ many genes. Hence each cell is represented by a $d$-dimensional vector $\bu$. The expression of each gene is quantified as a positive integer, and a higher value in a coordinate of a cell implies that the corresponding gene has high expression for that particular cell. 
Here we have two important inferences w.r.t to the clustering data from the preceding paragraph. 
Firstly that for each cell $\bu \in S$, only a few number of the coordinates have non-zero value.
Secondly, the ground-truth clusters that we are given, may have underlying sub-clusters (many of them could be even unknown by the research community). An important problem now is to identify the clusters of cell according to its gene expression.

\paragraph{Importance of clustering single-cell data} 
Single-cell analysis has received a huge number of attention in recent years. It has largely improved the study of many biological areas. For example, in recent years, more and more new target of cancer therapy and autoimmune disease therapy were discovered through this technique. And nowadays, it's widely used in almost all the life science research institutes. 
Yet there are still a lot of problems remained to be solved. 
For example, single-cell RNA sequencing data are extremely high dimensional, which may contains expression information of thousands genes from thousands, or even millions of cells. It's very challenging to achieve classification and clustering of these cells efficiently and precisely. Improvement of the computational method of single-cell analysis will benefit researchers globally, lead to more science breakthrough and save patients lives.
Lastly, we give a quote from Science's overview on \emph{Breakthrough of the Year 2018}\footnote{https://vis.sciencemag.org/breakthrough2018/finalists/}, it says,
`` \emph{The single-cell revolution is just starting.}''

\paragraph{Current clustering landscape.}Given the importance of single-cell RNA-seq data, single-cell analysis has become a highly active area. Many algorithms have been proposed and used in last several years. We refer \cite{kiselev2019challenges} for a good summary.
Besides introducing algorithms, many of works \cite{freytag2018comparison,duo2018systematic,luecken2019current} have made efforts to compare existing algorithms. These comparison results were obtained by testing existing algorithms on some public standard datasets. Seurat \cite{hao2021integrated} is one of the most popular algorithms in the biological research community, and it has also been observed accurate in many (but not all) datasets. In this paper, aside from the labels of \cite{zheng2017massively} associated with the dataset, we use performance of Seurat as a benchmark.

Against this background, we describe the concept of confident clustering.

\subsection{Confident clustering: motivation and accuracy}
In many single-cell data, there maybe some cells that are not clearly identifiable as part of being any one ground truth cluster $V_j$, but is rather a boundary/intermediate point.
In this paper, we use different subsets of the data provided by~\cite{zheng2017massively}. This is a popular benchmark dataset consisting of different cell types. There is a type of cell called \emph{hematopoietic stem cells} (HSC). In our dataset, all other cells are differentiated from HSC. In other words, HSC is the ancestor of other cells. When we ran Seurat on this dataset, Seurat and the labels provided by \cite{zheng2017massively} have a lot of conflicts on HSC and monocytes. Many cells which are identified as HSC by \cite{zheng2017massively} are identified as monocytes by Seurat. There are three possible reasons:
\begin{enumerate}
    \item Seurat made mistakes.
    \item \cite{zheng2017massively} made mistakes.
    \item Those conflicted cells have not fully differentiated, i.e., they are intermediate cells.
\end{enumerate}
As the biological data is complicated, it is hard to determine which case it is. In our algorithm, we address this issue by an alternative approach. 
We output two clustering results. 
In the first round, we only cluster 
cells that have high ``signal'' of membership to a cluster, which we denote as primary clustering. 
This generates a primary clustering on a large subset of the total number of cells. 
We observe that this result is highly \textbf{consistent with both Seurat and the labels obtained from~\cite{zheng2017massively}}. Meaning, our algorithm simply excludes cells that have mixed signals when forming the primary clusters. This further solidifies the possibility of ``intermediate/boundary'' cells. 

Once we have our primary, confident clustering,  
we cluster the rest of the points using the primary clustering as a backbone. 
\emph{For our dataset we verify that primary clustering has very low error w.r.t the ground truth clustering, even compared to the state of the art such as Seurat.} We further note that even though the misclassification 
increases in our second round of clustering, it remains comparable, (and in some case better) than state of the art.

\paragraph{Robustness of confident clustering}
Aside from accuracy, the other important characteristics of any practical clustering algorithm 
is its robustness. 
Almost all popular clustering algorithms (From fundamental algorithms such as K-means to more domain specific algorithms such as Seurat) have some user input parameters on which the outcome depends. It is desirable that the output of the algorithms are robust to small changes in these parameters, as long as they are within a reasonable range. However, it is hard to achieve in practice.

For example, many clustering algorithms, including us and Seurat, use PCA and projects the $d$-dimensional data to some $k'$ dimensional space, where $k'$ is an user input. Additionally, Seurat has a parameter ``resolution'', which decides the granularity of clustering, where higher resolution means Seurat outputs more clusters. For our dataset, for a fixed resolution=$0.5$, Seurat has moderately high error for $k'=10$, lower error for $k'=20$, and then again high error for $k'=30$. Such non-monotonous behaviour implies that the algorithm is sensitive to the choice of $k'$, and it is undesirable. In this regard we have the following observation.

\begin{observation}
Our confident clustering is robust to the choice of $k'$ within a considerably large range (between 10 to 30 for our dataset) with other parameters fixed. Furthermore, 
it is relatively unaffected by change in the other parameters, and we achieve high accuracy for a wide range of user inputs. 
\end{observation}

The accuracy and robustness of the confident clustering is the main motivation behind our algorithm. Boundary vectors seems to be a natural characteristics of vector clustering problems, one that can mislead algorithms not specifically designed to handle such case. 
In this paper we focus on single-cell sequencing data, but it will be interesting to observe similar phenomenons in other domains. Against this backdrop, we describe the organization of the rest of the paper.

\paragraph{Organization}
In Section~\ref{sec:2}, we give a high level idea of our philosophy and the motivation behind the steps. In Section~\ref{sec:3} we give a detailed description of our algorithm, highlighting outcome of different modules, their connection to theory and impact on the algorithm. Section~\ref{sec:4} concentrates on the experimental verification. We compare the accuracy of our algorithm w.r.t the labels provided by \cite{zheng2017massively} and that of Seurat's outcome in many natural setting, and also demonstrate the robustness of our algorithm. We finish in Section~\ref{sec:5} with a short summary and many future directions.

\section{Our approach on confident clustering\label{sec:2}}

Now we give an overview of our algorithmic framework. There are four main steps in our algorithm.
\begin{enumerate}
    \item Create an embedding of the vectors onto a graph so that most of the edges are between vertices representing vectors from same cluster.
    We will denote this as an \emph{approximate correlation graph.}
    
    \item Create many small disjoint sets such that each set is an almost subset of one of the clusters, having very small intersection with other clusters. We shall call these sets as \emph{confident sets.} 
    Note that these sets do not cover all vectors. We are left with some unclassified vectors. 
    
    \item Merge confident sets to form a smaller number of clusters, where each cluster consists almost entirely of one underlying true clusters. We call them \emph{primary clusters}. We hope 
    that a primary cluster has only very few errors, at the expense of  excluding ``remaining'' vectors of the previous step, which are still not classified. 
    
    \item Finally we add these unclassified vectors to the primary clusters using a simple absolute majority voting to complete our clustering. 
    
\end{enumerate}

In the following part of this section, we explain the intuition and the terminology behind each step. 
We start with recalling the notion of PCA's 
compressibility.

\paragraph{PCA compression.}
Let $\hat{A}$ be the matrix of datapoints as well explained before. We first recall the definition of PCA compression ratio discussed by Mukherjee and Zhang \cite{mukherjee2022compressibility}.
\begin{definition}[Compression ratio of PCA]
Let $\bu$ and $\bv$ be any two vectors in the data set
$\hA$ and $P^{k'}_{\hA}(\bu)$ and $P^{k'}_{\hA}(\bv)$ be their
projection due to the top $k'$ Principle components. 
We define the $k'$-PC compression of two vectors $u$ and $v$
as the ratio
\[
\Delta_{\hA,k'}(\bu,\bv)=\frac{\|\bu-\bv\|}{\|P^{k'}_{\hA}(\bu)-P^{k'}_{\hA}(\bv)\|}
\]
\end{definition}

Mukherjee and Zhang observed that, in a generic theoretical as well as many application settings, PCA projection compresses the $\ell_2$-norm distance of intra-cluster pairs. That is, $\Delta_{\hA,k}(\bu,\bv)$ is significantly larger than $\Delta_{\hA,k}(\bu',\bv')$ if $\bu$ and $\bv$ are from the same cluster and $\bu'$ and $\bv'$ are from different clusters\footnote{This is not true for all pairs $(\bu,\bv)$ and $(\bu',\bv')$. But it is true for many typical pairs.}. In particular, this phenomena has been observed in single-cell RNA sequencing data. 
Given this observation, we know that the higher the compression ratio of a pair, 
more is the likelihood that they are from the same cluster. 
Now we define a correlation graph, that allows us to use PCA's compressibility 
in a graph-embedding setting.

\paragraph{Correlation graph.}
Let there be a set $S$, and $V=\{V_1,\dots,V_k\}$
be the ground truth clustering, we define the graph correlation $G$ as follows:
\begin{definition}[Correlation graph]
Let $V=\{V_1, \ldots ,V_k\}$ be the ground truth 
clustering of $S$. 
The associated correlation graph is a graph $G_{S,V}$ with $|S|$ nodes so that 
for each pair of nodes $\bv,\bv'\in S$, there is an edge connecting them if and only if the they are from a same set $V_{\ell}$, where $\ell\in [r].$ 
\end{definition}

Thus, the correlation graph $G_{S,V}$ is a certificate of this clustering, 
one that allows clustering with $100\%$ accuracy. 
However, creating such a graph from $S$ is a bleak prospect. 
Instead, we focus on an ``approximate correlation graph'', 
that allows for some deviation from the absolute definition. 

\begin{definition}[$(\alpha,\beta)$-correlation graph]
Given a set $S$ and a ground clustering $V$, a graph $G_{S,V}^{\alpha,\beta}$  is called a $(\alpha,\beta)$-correlation graph if it has $\alpha$-fraction of the ``intra''-cluster edges, i.e., $\alpha$ percent of edges in $G_{S,V}$ also appear in $G_{S,V}^{\alpha,\beta}$
and $\beta$-fraction of the ``inter''-cluster edges, i.e., only $\beta$-fraction edges of $G_{S,V}^{\alpha,\beta}$ are not edges in $G_{S,V}$.

Thus a correlation graph can be described as a $(1,0)$-correlation graph.

\end{definition}

In our first step of the algorithm, we hope to construct an approximate correlation graph with large $\alpha$ and low $\beta$. Such an approximate correlation graph enables us to 
obtain a clustering that is very close to $C$, the ground truth cluster using
graph clustering techniques. 
We create this graph using the observation of Mukherjee and Zhang regarding compressibility of PCA~\cite{mukherjee2022compressibility}, which they verified the in both random vector model (theory) and single-cell datasets.

In brief, we define a vertex corresponding to each vector, and connect 
two vertices if the compression ratio due to PCA for the pair is in 
the top $5\%$ among all pairs.
This step gives us a correlation graph $G^{\alpha,\beta}_{S,C}$ 
where $\alpha>>\beta$. This big gap between $\alpha$ and $\beta$ is crucial for our next step.

Before we describe our next step, we define the concept of confident set. 

\begin{definition}[$\zeta$-confident set]
Given a set of vectors $S$ and ground truth clustering $V=\{V_1, \ldots V_k\}$,
a set $U$ is a $\zeta$-confident set if 
There exists $i$ such that $|U \cap V_i| \ge \zeta|U|$
\end{definition}

In the second step of our algorithm, we hope to recover large-size and high $\zeta$ confident sets from $G^{\alpha,\beta}_{S,C}$. 
From hereon, we denote $G^{\alpha,\beta}_{S,C}$ with simply $G$. 
In this step, we aim to find large cliques (or highly dense subgraphs) in $G$. We use these densely connected graphs to capture confident sets. Since $G$ approximates the correlation graph, the nodes in a reasonably sized densely  clique of $G$ should has high chance to be in the same cluster. Our method to find large cliques here is motivated from 
graph generative models, especially the random graph model and stochastic block model (SBM).
Here we give a high level description of our idea. 
Roughly speaking, we observe, both in probabilistic theoretical model and in 
single cell data, the eigenvectors of the adjacency matrix of $G$ correspond
to very highly connected sets of vertices. Coupled with the fact that $G$
is an $(\alpha,\beta)$-correlation graph where $\alpha>>\beta$, this allows us to recover many
substantially large confident sets from $G$ via a simple spectral algorithm that we device.

At the end of this step, we have many confident sets, and a set of unclassified vertices that are not included in any confident set. 
In our experiments, our confident sets have very high accuracy, against several choice of the parameters that we describe in Section \ref{sec:4}.  A detailed algorithm will be discussed in Section \ref{sec:3-recover}.

In the next step, we merge these sets to form the primary clusters, 
which are large clusters, with their number close to the number 
of underlying clusters. Essentially, a primary cluster can be thought as a union of several confident sets. Furthermore, its size is comparable to the ground truth cluster which constitutes its majority.

We form these clusters by merging the confident sets using a greedy algorithm based on the neighbourhood of each vertex w.r.t to the compression ratio metric. 
For our dataset, we have around 25-25 confident sets depending o the choice of $\gamma$. After our merge procedure, we are always able to recover $7$ low error primary clusters. We define this step formally in Section~\ref{sec:3-merge}.

At this stage, we have our first clustering result. Here two things are important to note. 

\begin{enumerate}

    \item We have clustered the cells about which we were highly confident in our previous steps. The aim of this philosophy is to cluster a large subset of the cells with minimal errors.
    This also has the possible implication that the cells that we do not cluster in this round, 
    are more likely to be boundary cells. We discuss this in detail in the later sections. 
    
    \item Once we have the primary clusters, we can cluster the ``remaining'' vectors
    using a very simple majority voting. We believe these are cells that showed comparatively 
    weak signals in our steps. This step does increase the error in the dataset, but we remain comparable to the state of the art. 

\end{enumerate}

\section{Analysis of the main algorithm} 
\label{sec:3}

In this section, we describe the different steps of our algorithm in detail, 
mainly focusing on experimental data to discuss its reasoning. As discussed
above, our approach is motivated by PCA's proof of relative compression in a 
probabilistic setup~\cite{mukherjee2022compressibility} and success of 
spectral algorithms in graph generative models such as the stochastic block model~\cite{mcsherry2001spectral}. 
We believe our algorithm can be analyzed in a generic random vector model. 
Similar analysis have been studied in many papers such as \cite{mcsherry2001spectral,Vu2007,mazumdar2017clustering, peng2021towards, mukherjee2022recovering,mukherjee2022compressibility}.

Even though we focus on experimental results in this paper, we give brief arguments supporting the sensibility of our algorithm in probabilistic models. 
We first discuss the datasets of our interest.

\subsection{An explanation of the testing dataset \label{sec:3-dataset}}

In our experiments, we focus on the dataset provided by \cite{zheng2017massively}. This dataset is avaible at the website of 10xGenomics\footnote{https://www.10xgenomics.com/resources/datasets}. This dataset is a combination of \emph{immune cells}. We first explain the cells.

\begin{figure}[H]
    \begin{center}
    \includegraphics[scale=0.5]{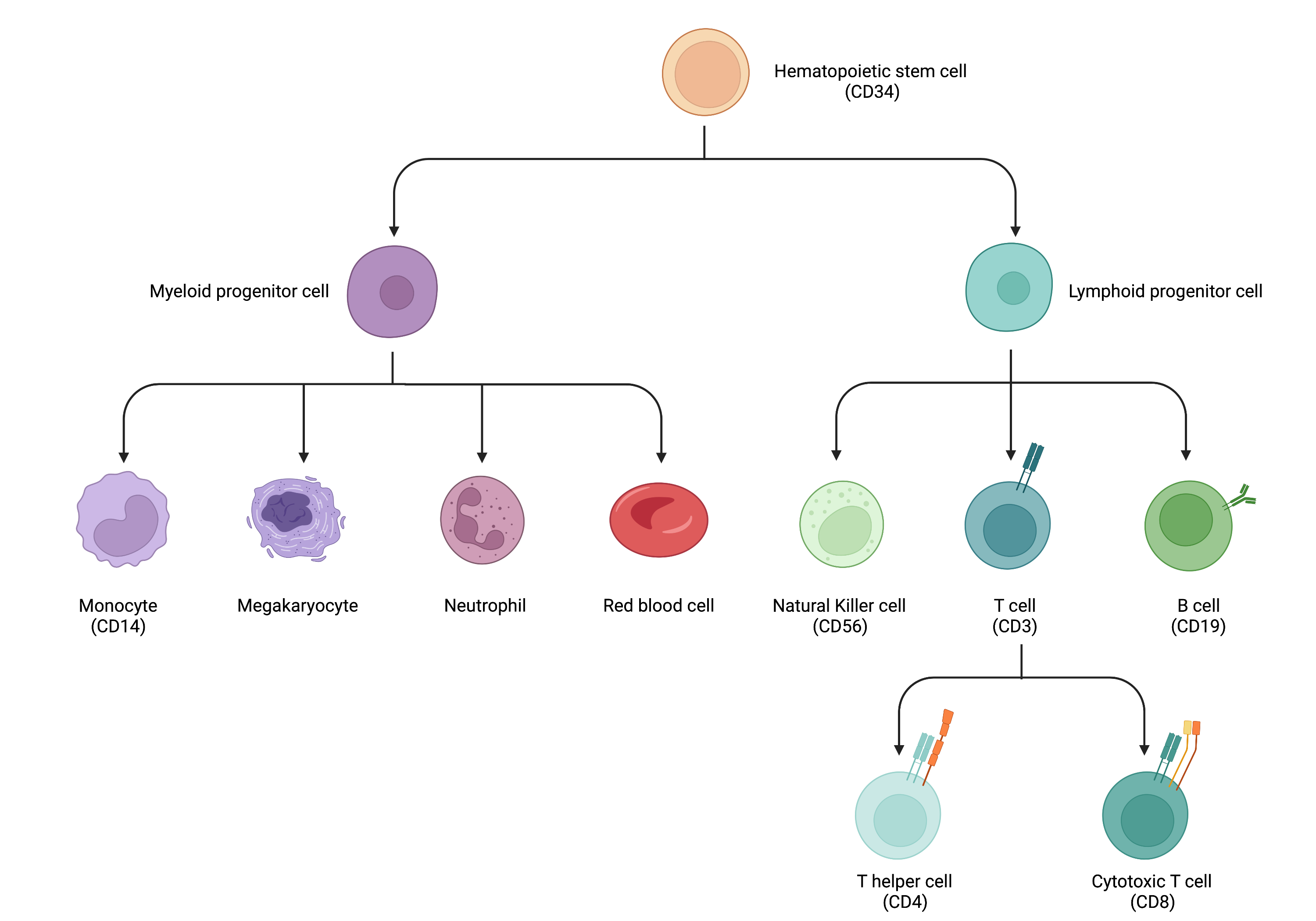}
    \end{center}
    \caption{Immune cell lineage and markers}
    \label{graph:immune_cells}
\end{figure}

We use $6$ clusters from this dataset, with the original number of cells in them being  (11213, 10209, 2612, 10085, 9232, 8385). 
The corresponding clusters are CD4 T cells, CD8 T cells, CD14 monocytes, CD19 B cells, CD34 HSC and CD56 natural killer cells respectively.

The label of the cells in \cite{zheng2017massively} are decided on the basis of expression of a particular gene in them. For example, a cell is labeled as CD34 if it has CD34 expression or related population marker genes. In biological meaning, these cells are very likely to be hematopoietic stem cells.
However, because of complexity of biological data, the labeling still has a tiny chance to be wrong. 
Similarly, cells in the CD4 cluster are very likely to be T-helper cells.

Furthermore, it is important to note that these cells are not fully separable. As discussed before, hematopoietic stem cells (CD34) are the common ancestor of all other five types of cells.
In fact Seurat and \cite{zheng2017massively} have a lot of conflicts between CD34 and CD14. 
Also, T helper cells (CD4) and cytotoxic T cells (CD8) are both T cells, it is technically harder to distinguish them computationally compared to other cells. It has also been discussed by \cite{hao2021integrated}. 
More biological connections between these cells are in Figure \ref{graph:immune_cells}.

Thus, even though there may be a small number of errors, the labels \cite{zheng2017massively} are still trustable. Hence we use it as a benchmark in our comparison.

\subsubsection{Our datasets}
For our experiments, we choose each cell belonging to these $6$ clusters with probability $\frac{1}{10}$. 
In this way, we create two datasets. 

\begin{enumerate}
    \item Dataset-1: This dataset has a total of 4743 cells, with 
    $(1022,924,260,955,839,743)$ many cells with corresponding to the clusters CD4, CD8, CD14, CD19, CD34 and CD56 respectively. 
    \item Dataset-2: This dataset has a total of 4780 cells, with 
    $(1035,893,258,914,893,787) $many cells with corresponding to the clusters CD4, CD8, CD14, CD19, CD34 and CD56 respectively.
    
\end{enumerate}

In this paper, we describe our experiments for Dataset-1. 
However, we obtain similar results for Dataset-2. 

Before we give a step-wise description of our algorithm, we describe the setup of Seurat on a high level. This allows us to make more in depth comparisons between Seurat and our algorithm.
\paragraph{High level sketch of Seurat}
 The basic workflow of Seurat is as follows.
\begin{itemize}
    \item Run a supervised PCA 
    \item Create a $k$-NN graph according to post-PCA distance
    \item Apply a graph clustering algorithm (Louvain \cite{blondel2008fast}) on the $k$-NN graph
\end{itemize}

Now we describe our algorithm step by step. 

\subsection{Constructing an approximate correlation graph via PCA compression ratio
\label{sec:3-corr}}

We first describe our algorithm about the construction of approximate correlation graphs.

\begin{algorithm}[!htb]
\caption{\textsc{Construction of Correlation Graphs}}\label{alg:corr_graph}
% \KwData{$G=(V,E),p,q,k$}
% \KwResult{Clusters of size at least  $\frac{n}{1.1k}$}
% \SetKwFunction{FMain}{\textsc{Cluster}}
% \SetKwProg{Fn}{Function}{:}{}
%   \Fn{\FMain{$G=(V,E),p,q,k$}}{
\begin{algorithmic}[1]
%\Procedure{\textsc{Cluster}} {$G=(V,E),p,q,k$}
\STATE Takes as input $\hA\in\mathbb{R}^{d\times n}$.
\STATE Fix $\gamma$ and $k'$ 
\FOR{each pair of columns $\bu,\bv$ of $\hA$}
 \STATE Compute $\Delta(\bu,\bv)\gets \|\bu-\bv\|/\|P^{k'}_{\hA}(\bu)-P^{k'}_{\hA}(\bv)\| $
\ENDFOR
%\STATE 
\STATE Sort all pairs of $(\bu,\bv)$ according to the non-increasing order of $\Delta(\bu,\bv)$
\STATE Let $B$ be the top-$\gamma$ percent of pairs in the sorted array
\STATE Construct a graph $G$ with $n$ nodes each of them represent a column $\bu$;
\STATE Add an edge between $\bu$ and $\bv$ if and only if $(\bu,\bv)\in B$
\STATE Output $G$
\end{algorithmic}
\end{algorithm}

Here we first project the data onto a $k'$ dimensional space using PCA. Then we embed this projected data onto 
a $n$-vertex graph. Each vertex corresponds to a cell, 
and we add an edge between two vertices if their $k'$-$PC$ compression ratio lies within the top $\gamma \%$ of all the pairs. This gives us our approximate correlation graph $G$.
In this paper, we typically choose $\gamma$ between $3\%$ and $5\%$ and $k'$ varies from $10$ to $30$.

\paragraph{Theoretical background}
Let us now explain the theoretical rationale and experimental evidence behind this step. As we have discussed, an 
$(\alpha,\beta)$-correlation graph with $\alpha>>\beta$
implies most of the edges in the graph are intra-cluster
edges w.r.t to the ground truth clustering. 

In this direction, we discuss the compressibility result by Mukherjee and Zhang~\cite{mukherjee2022compressibility}
in brief before moving ahead. Roughly speaking, \cite{mukherjee2022compressibility} showed that, $\Delta(\bu, \bv)$ is likely to be larger if $(\bu,\bv)$ is an intra-cluster pair.

The theoretical result of~\cite{mukherjee2022compressibility} is based on a random (stochastic) vector model. In this setup the dataset consists of 
$n$ many $d$-dimensional vectors $\bui \in R^d$, which is represented
as a $d \times n$ matrix $\hA$. There are $k$ many ground truth clusters $V=\{V_1, \ldots , V_k\}$, each  cluster $V_j$ associated with a center $\mat{c_j} \in R^d$. Each cluster is also associated with a $0$-mean distribution $\mD^{(j)}$. The only distributional assumption
on $\mD^{(j)}$ is that it is coordinate wise independent. 
The dataset is then defined as follows. 
If a vector $\bui$ belongs to the $i$-th cluster 
then $\bui$ is sampled as $\bui=\mat{c_i}+X_i$ where $X_i \sim \mD^{(j)}$. 
That is, the vectors can be considered a summation of their underlying centers and a random vector that is sampled from this distribution
$\mD^{(j)}$.
Here $V, k$ and $\mD^{(j)}, 1 \le j \le k $ are unknown. Informally speaking, they have the following result.

\begin{theorem}[Informal version of \cite{mukherjee2022compressibility}]
\label{thm: compress}
Let's say $\hA$ is defined according to the aforementioned random vector model, 
such that in the dataset the distance between intra-cluster and inter-cluster 
points are very similar.
Then for an appropriate choice of $k'$, the $k'$-PC compression ratio of
intra-cluster points is much higher than that of inter-cluster points. 
\end{theorem}

While this model is fairly general, as it does not make any distributional assumption, it is still simple in comparison to reality. However, they also complemented this result with observation on real datasets, showcasing this phenomenon for for two benchmark datasets. In this direction we also verify these statistics for our arguably more complex dataset-1. For a precise statement of Theorem \ref{thm: compress}, we refer to the paper \cite{mukherjee2022compressibility}.

Thus, the choice of $k'$ and $\gamma$ effects the output of the algorithm. The work in \cite{mukherjee2022compressibility} demonstrated that in the 
random vector model, the best choice of 
$k'$ is the number of ground truth clusters. However, the ground truth is unknown, and thus we do not know the number. Fortunately, \cite{mukherjee2022compressibility} also argued that the choice of $k'$ is not that sensitive, and a number in the proximity of true number of clusters still gives us good compression ratio. This has also been observed in practice.

\begin{figure}[h]
    \centering
    \includegraphics[width=10cm, height=4.5cm]{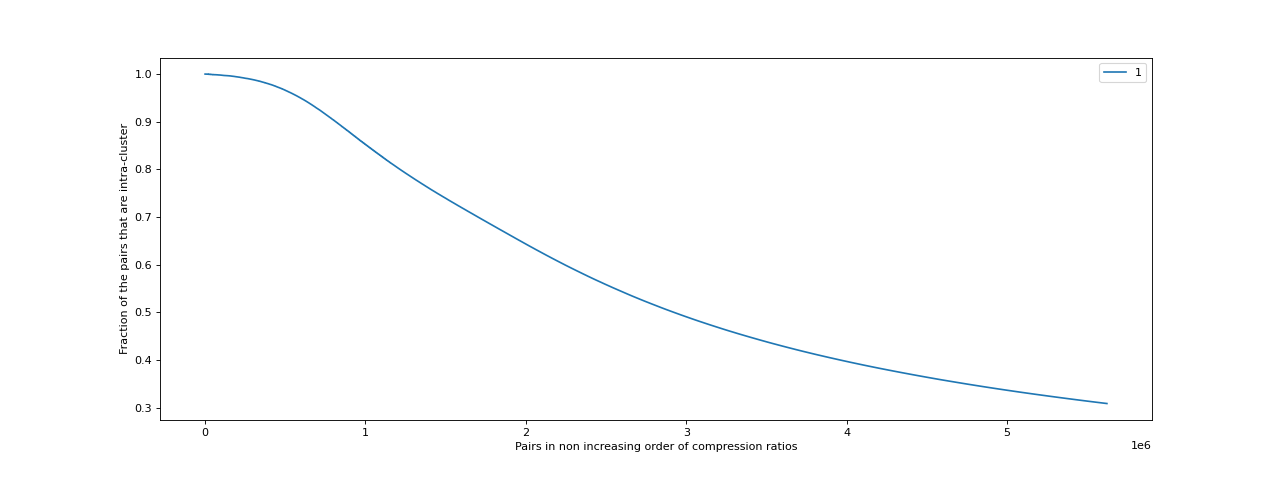}
    \caption{$6$-cluster graph}
    \label{fig:corrfraph-dataset3}
\end{figure}

Secondly, the choice of $\gamma$ effects the value of $\alpha$ and $\beta$. 
If we choose $\gamma$ to be very small (say less than $1\%$, the graph will be too sparse,
and we won't be able to apply spectral techniques that work in the ``dense graph'' regime.
On the other hand, 
\cite{mukherjee2022compressibility} observed that for two benchmark datasets, among the top $10\%$ of the edges, $99\%$ were intra-cluster. This
also holds for the arguably more complex dataset-1, which is our center of 
interest in this paper. We now describe the experimental evidence of the choice of 
$\gamma$ on $\alpha$ and $\beta$ for dataset-1 in Figure\ref{fig:corrfraph-dataset3}.
Here $x$-axis denotes the fraction of total pairs (we only mark the top $60\%$) 
and the $y$-axis denotes the corresponding fraction of those points that are intra-cluster.

For example, in Figure\ref{fig:corrfraph-dataset3}, more than $98\%$ pairs in top-$5\%$ compressed pairs are intra-cluster pairs. Meaning $\beta$ is smaller than $0.02$. 
In Section~\ref{sec:4-robust} we show that our algorithm works for a lower choice of $\gamma$ as well.
This way we form our approximate correlation graph $G$. Once we have this graph, 
we then recover many disjoint confident sets from the graph.

\subsection{Recover confident sets from the approximate correlation graphs
\label{sec:3-recover}}
Given an approximated correlation graph $G$, our next step is to find the confident sets. Recall that a $\zeta$-confident set is one that has more than $\zeta$-fraction of points (vertices) belonging to any one ground truth cluster. 
Now, since we work on this embedded graph $G$, the problem resembles a graph clustering problem. We note that Seurat also used this strategy. 
However, there are several important differences.

Seurat creates the graph $G'$ by connecting every vertex to their closest $20$ neighbours based on post-PCA distance, and then runs Louvain algorithm on it, which is a popular graph clustering algorithm. This has the following implication. 

\begin{enumerate}
    \item Seurat creates a regular ``sparse'' graph,
    based on a $k$-NN strategy.
    In comparison we create a dense and irregular graph. For example, in the graph created by choosing top $5\%$ edges for dataset-1, we have multiple cliques 
    of size $300$. 
    
    \item Although Louvain is widely used in practice, there are only a few number of theoretical analysis of algorithms \cite{cohen2020power}.  We use spectral techniques to extract confident sets out of our dense graph, and these techniques are not applicable to sparse graph.   

\end{enumerate}

\begin{algorithm}[!htb]
\caption{\textsc{Finding confident sets}}
\label{alg:confident-set}
% \KwData{$G=(V,E),p,q,k$}
% \KwResult{Clusters of size at least  $\frac{n}{1.1k}$}
% \SetKwFunction{FMain}{\textsc{Cluster}}
% \SetKwProg{Fn}{Function}{:}{}
%   \Fn{\FMain{$G=(V,E),p,q,k$}}{
\begin{algorithmic}[1]
%\Procedure{\textsc{Cluster}} {$G=(V,E),p,q,k$}
\STATE Takes as input the graph $G$
from Algorithm~\ref{alg:corr_graph},
expressed as the adjacency matrix $A^{G}$.
\STATE $G' \gets G$, $i \gets 0$
\REPEAT
\STATE  Obtain $\hat{v}$ $\gets$ the highest eigenvector of $A^{G'}$ \hfill[$G'$ is the residual graph]
\STATE Sort the columns $\bui$ of $A^G$ in the non-decreasing order of 
$\langle \bui,\hat{v}\rangle$
\STATE Select the top $c$ vertices form the list: $U_c$, 
s.t \\
$c \gets \max\{ t: \text{There are a total of $\frac{1}{2}{{t \choose 2}}$  edges between the top $t$ vertices}  \}$
\STATE $avg \gets \frac{c}{2}$
\STATE $G'' \gets G'(U_c)$, \hfill [$G''$ the induced graph due to $U_c$]
\STATE $CS_i \gets \{\bu: \bu \in U_c, \deg(\bu) \text{ in } G'' >avg\}$
\hfill [Remove low degree vertices]
%\STATE $CS_i \gets U_c$
\STATE $G' \gets G' \setminus CS_i$ \hfill [Remove the vertices in $CS_i$ ]
\STATE $i \gets i+1$
\UNTIL{$U_c<(\frac{2}{3}\gamma*100)$}
\STATE $R \gets  S \setminus
\left(  CS_0 \cup \ldots \cup CS_{i-1}  \right)$
\STATE Output $CS_0, \ldots ,CS_{i-1}$, $R $ 
%	\EndProcedure
%}
%\end{algorithm2e} 
\end{algorithmic}
\end{algorithm} 

\begin{figure}[h]

    \begin{center}
    \includegraphics[scale=0.31]{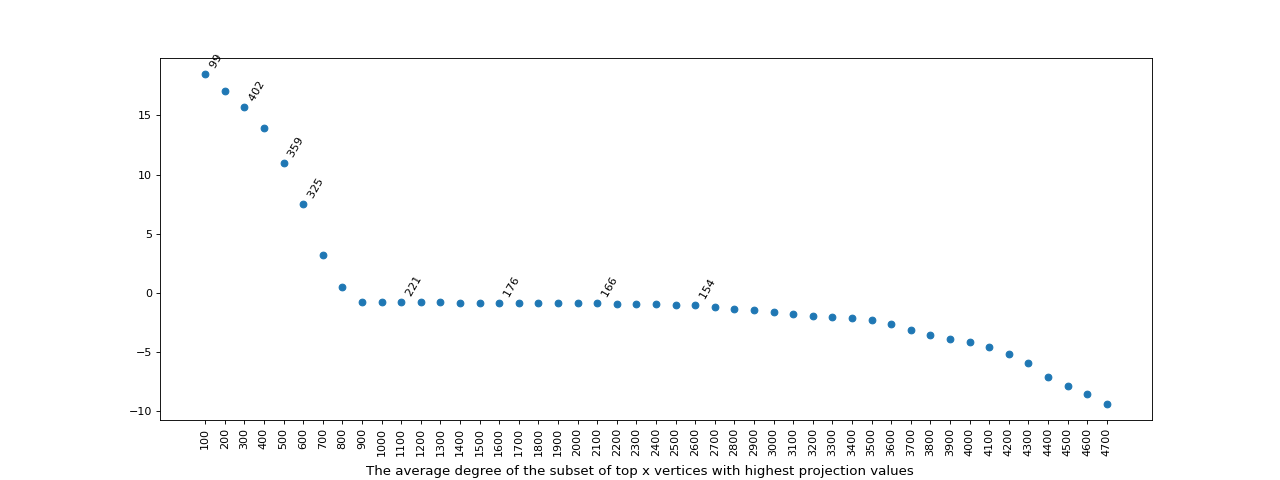}
    \subcaption{Sorted projection values with first PC}
    \includegraphics[scale=0.31]{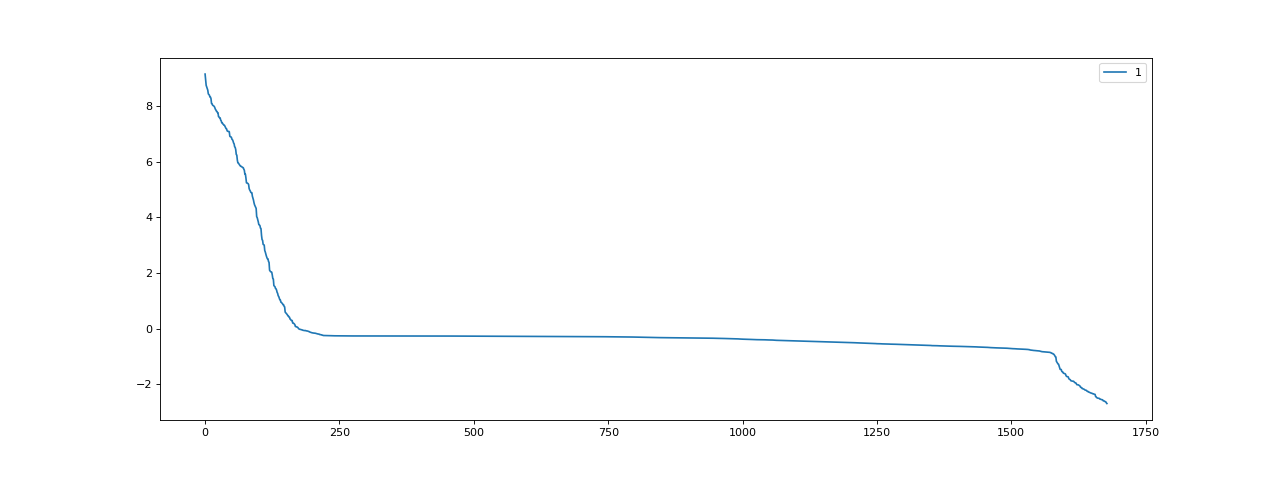}
    \subcaption{Sorted projection values with tenth PC}
    \end{center}
    \caption{Projection value and degree for first and tenth PC}
    \label{fig:sproj10}

\end{figure}

At every step, we aim to find a dense subgraph in current residual graph $G'$. We obtain it by taking the top eigenvector $\hat{v}$ of the adjacent matrix $A^{G'}$, and then take the vertices whose columns have the highest correlation with $\hat{v}$, checking that that they form a densely connected component. The idea is inspired by spectral algorithms for planted graph problem and SBM. Unlike the clean theoretical model, we also add some steps in line 4 - line 9 of Algorithm \ref{alg:confident-set} to remove some noise vertices, which are the vertices with comparably low degree in that subset.
We experimentally verify that the top vertices (vertices with highest projection w.r.t to an eigen vector) 
do form densely connected subgraphs in Figure~\ref{fig:sproj10}, for different eigenvectors. 
Here in $x$-axis we have the vertices in the graph sorted according to their projection value,
and in $y$-axis we have the projection value.
As we can observe, for both the eigenvectors, the projection value is $0$ for most of the vertices. If we look at the vertices that have larger than $0$ projection, they are likely to be densely connected (we describe their average degree for the first eigenvector). 

We believe this can also be proved in a random graph setting, which is an interesting question to explore.

We choose a threshold such that 
the average degree of the induces graph due to the vertices is half of the size of the set. Recall that if we do find a densely connected component, then it has high chance of being from the same ground truth cluster if our approximate correlation graph has
good $\alpha$ and $\beta$ value.

\paragraph{Theoretical background behind Algorithm~\ref{alg:confident-set}}
The rationales behind using the top eigenvector based approach are as follows. Eigenvectors and eigenvalue spectrum are important tools in the analysis and recovery
of densely connected subgraphs, components of stochastic block models and hidden cliques. 

In the stochastic block model, there is a graph with underlying cluster identity, such that if two vertex belong to the same clusters, there is an edge between them with probability $p$, otherwise there is an edge with probability $q$, where $p>q$. 
In this setup, we have the following fact

\begin{fact}
In the mean adjacency matrix for an SBM graph, the eigenvectors 
each correspond to a cluster.
That is, for each eigenvector, the inner product with the vertices belonging to
the cluster has highest value.

Here the mean adjacency matrix $A'$ is formed by putting $A'_{i,j}=p$ if 
$i$ and $j$ belong to the same cluster, and $q$ otherwise. 
\end{fact}
This behavior also flows over to the sampled SBM graph in a more general sense,
captured in the following informal theorem.

\begin{theorem}[\cite{mcsherry2001spectral,Vu2007,mukherjee2022recovering,Abbe2SBM}]
In the stochastic block model as well as in the random vector model with $k$ underlying clusters, the top $k$ eigenvectors 
correspond to the $k$ clusters, so that when the vectors are projected onto the subspace of the top $k$ eigenvectors, 
vertices (vectors) from 

Especially, Abbe et. al.~\cite{Abbe2SBM} show that if there are only two clusters in the SBM model, the top two eigenvectors of the centered adjacency matrix each has positive valued entry for rows representing vertices from one cluster, and 
negative values entry for other clusters. 
\end{theorem}

Similarly, in the hidden dense subgraph problem, there is a random graph where any pair of vertices are connected with probability $1/2$, and then a dense subgraph is planted on this random graph. The hidden clique problem is the case where the dense subgraph is in fact a clique. It has been shown that in this case the first eigenvector of a slightly modified (centered) adjacency matrix can recover this planted clique~\cite{AKS-spectral}. We believe an analysis similar to this can explain
why the vertices of our approximate correlation graph with high correlation with an eigenvector forms a dense subgraph, as witnessed in~\ref{fig:sproj10}.

The correlation graph has a structure that can be something intermediate between 
the stochastic block model and hidden dense subgraph problems, as $G^{\alpha,\beta}_{S,V}$ has several dense subgraphs for each of the underlying clusters, but there are also vertices with relatively low degree. We believe the analysis for SBM and hidden dense subgraph model can also be applied to our graph structure in stochastic settings similar to~\cite{mukherjee2022compressibility}. We verify this step with the following experimental evidence.

\paragraph{Experimental evidence}

For dataset-1, working with the threshold generates $23$ sets. Firstly we note that Algorithm~\ref{alg:confident-set} allows to get many large dense subgraphs. The first $5$ sets $U_c$ we recover are of size $630, 562, 450, 440$ and $245$. 
The size of the ground truth clusters corresponding to the majority vertices (always more than $90\%$ in our experiments) are between $700$ to $1000$. This shows that
\begin{itemize}
    \item The approximate correlation graph $G^{\alpha,\beta}_{S,V}$ we create contains many large densely connected components of vectors with same ground truth clusters.
    
    \item The spectral technique in Algorithm~\ref{alg:confident-set} is capable of extracting many such components.  
\end{itemize}

Then, once we filter these dense subgraphs to only keep the highly connected vertices, their performance as 
confident sets are as follows.
\begin{enumerate}
    \item $\zeta=1$ for $13$  out of the $23$ sets. That is we recover $13$ sets with zero error.
    
    \item $1>\zeta>0.98$ for $5$ other sets.
    
    \item $\zeta>0.9$ for the other sets.
\end{enumerate}

Thus, we verify our intuition and analysis behind this 
step by comparing with dataset-1. Finally we discuss our stopping criterion.

\paragraph{Stopping criterion}
In this setup we have chosen the threshold for termination as the point when the confident set obtained has size less than some value $T$, where $T=c \times \gamma$. where $\gamma$ which is the percentage of vertex pairs that we connect to form our approximate correlation graph. 
The reason for relating $T$ with $\gamma$ is as follows. 
When we create the confident set, we look for densely connected subgraphs. If we reduce the number of edges present in the graph, 
then the size of confident sets will decrease, while causing an in crease in the number of them. 
\begin{remark}
Furthermore, it should be noted that if $T$ is very small, then finding densely connected subgraph of size $T$ does not make much sense. 
For example, even in a random graph on $n$ vertices (where any two vertices are connected with probability $1/2$), one can find a clique of size $\log n$ with high probability.  
\end{remark}
We have verified that our algorithm works for different choice of $\gamma$ and $k'$ for this particular constant. However, we believe a more theoretical analysis of the step will enable us get a more interpretable stopping criterion.

Now, if the size of the smallest ground truth cluster is larger than $T$, then we expect there to be some confident sets corresponding to each cluster. Otherwise, we can continue to our next step and finish our first round of clustering. Then we can run our algorithm just on the unclustered vectors at that point. We shall argue that even if there are confident sets extracted corresponding to a small cluster, it does not cause it to be misclassified in the downstream analysis, because of our conservative approach.

\subsection{Combine confident sets according to cut-edge density to form primary clustering}
\label{sec:3-merge}

In this step we merge the confident sets to obtain our primary clustering. We want to obtain the following properties:
\begin{enumerate}
    \item Each primary cluster has a high majority for one of the ground truth cluster. 
    
    \item Corresponding to any ground truth cluster, there are few primary clusters.
\end{enumerate}
Thus we want to construct $\hat{k}$ many high accuracy cluster where $\hat{k} $ is close to the number of true underlying clusters.
Now recall that in the a$(\alpha,\beta)$-correlation graph $G$, not all vertices have the same degree, as we are interested in only using the highest quality edges in terms of the compression ratio. Essentially, to keep $\beta$ very low, 
we need to choose a small percentage of edges, which results in a small $\alpha$ (although $\alpha>>\beta$). 
Thus the following scenario is very likely. 
Let $CS_i$ and $CS_j$ be two highly confident sets, both belonging to the same underlying cluster $V_{\ell}$. However in $G$, there might not be a very high number of edges going from $CS_i$ to $CS_j$. This can be a problem if we want to merge the confident sets based on $G$. Thus we take the following strategy. 

\begin{algorithm}[!htb]
\caption{\textsc{Forming the primary clusters}}\label{alg:confident-cluster}
\begin{algorithmic}[1]
\STATE Take as input $S$ and confident sets $CS=\{CS_1, \ldots CS_r\}$
\STATE Choose $\delta$
\STATE For every vertex $\bu \in S$, form 
$N_{\bu}^{\delta} \gets\{\bv: \Delta(\bv,\bu)\text{ is top }\delta\%\text{ among all pairs with } \bu\}$
%of the edges in terms of compression ratio with an end point in $\bu$.
\REPEAT
\FOR{$1 \le i,j \le r, i \ne j$}
\STATE $Y_{i,j} \gets \frac{1}{|CS_i|}\sum\limits_{\bu \in CS_i} \left| N_{\bu}^{\delta} \cap CS_j \right|$ \hfill [Note that this is asymmetric]
\ENDFOR
\FOR{$1 \le i,j \le r, i \ne j$}
\STATE $Z_{i,j} \gets Y_{i,j} \cdot Y_{j,i} $ \hfill [The score determines how to merge]
\ENDFOR
%\STATE Sort the tuples $\left(Z_{i,j},i,j  \right)$ in the non-increasing order of $Z$
\STATE $CS'_1\gets CS_i\cup CS_j$, where $CS_i$ and $CS_j$ has the maximum $Z_{i,j}$
\STATE $CS \gets CS'_1 \cup \{CS_{\ell}\}_{\ell \ne i,j}$
\UNTIL{User interrupt}
\STATE $PC \gets CS$
\STATE Output $PC$
\end{algorithmic}
\end{algorithm} 

To get around the irregularity of $G$, we do the following. 
For every vertex, $\bu$, we form a an array $N_{\bu}$ which consists of all the vertices in $S \setminus \{\bu\}$. 
This array is sorted in the non-decreasing order of 
$\Delta(\bu,\bv)$, the $k'$-PC compression ratio. 
Then we define $N_{bu}^{\delta}$ as the first $\delta \%$
of the vertices. That is, $N_{bu}^{\delta}$ is the set of 
the vertices for which $\bu$ has the highest compression ratio. For a small $\delta$, we hope that all these vertices belong to the same cluster as $\bu$.

Let us call $N_{\bu}^{\delta}$ as the close neighbours of $\bu$.
Then for any $CS_i, CS_j$ we calculate
$Y_{i,j}$, which is the average number of close neighbours the vertices in $CS_i$
have in $CS_j$. Similarly we calculate $Y_{j,i}$. This is similar to the $k$-NN graph used by Seurat, but crucially, we use this setup only \emph{after} we have already 
formed the confident sets. Then we we recursively merge the confident sets 
based on highest $(Y_{i,j} \times Y_{j,i})$ value. Here the reason for taking a product
of this term is to ensure that both $CS_i$ and $CS_j$ have a large close-neighbourhood in each other. We continue this step until we hit our stopping criterion (which is an user input in this case). Now we  discuss this step, along with the stopping criterion with some experimental outcome. 

\paragraph{Experimental outcome}

\begin{figure}[H]
    \centering
    \includegraphics[scale=0.34]{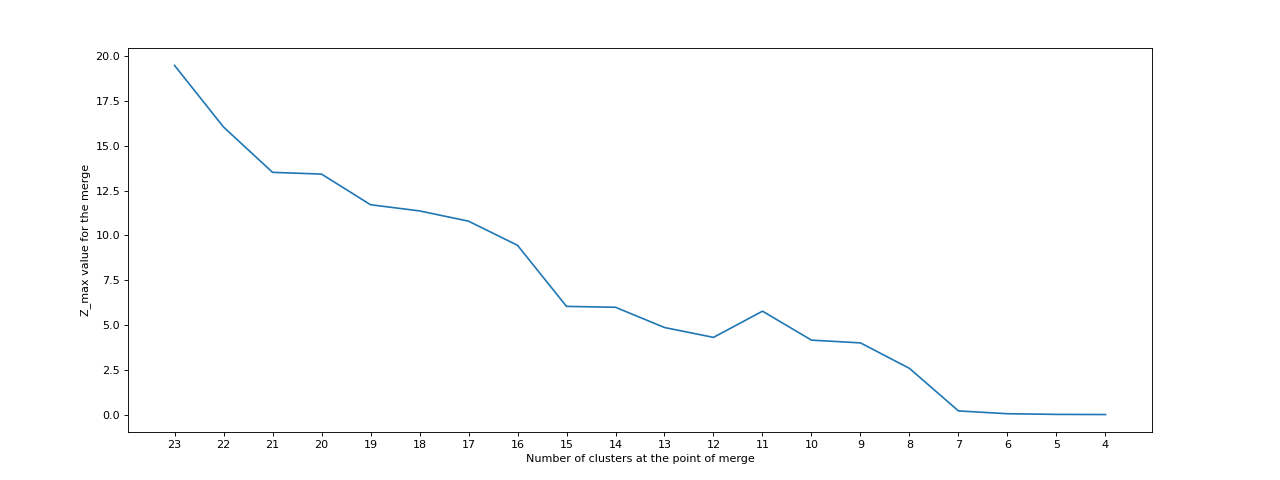}
    \caption{$Z_{\max}$ scores for $\gamma=0.05, k'=20$.}
    \label{fig:z_max}
\end{figure}

Recall that for dataset-1, we have $6$ underlying ground truth clusters, $C_1, \ldots C_6$ with their sizes being $(1022,924,260,955,839,743)$. 
Our algorithm for the current stopping criterion with $\gamma=0.05, k'=20$ and $\delta=2.5$ gives us $6$ primary clusters $PC_1, \ldots PC_6$. Here the $Z_{\max}$ value for the final merge is comparable to the previous merges, but the next merge (one that would form create $5$ clusters) is much lower, which we treat as a stopping point. We refer to the $Z_{\max}$ values for the merges as a metric, described in Figure~\ref{fig:z_max}.
for this graph, the $Z_{\max}$ values when the number of clusters post merge is going to be $7,6,5,4$, are $2.86, 2.57, 0.21, 0.045$ respectively. This shows a sharp fall in the value when we try to form $5$ clusters from $6$ clusters. The values for similarly for other values o $\gamma$ and $\delta$ at the point where a merge would cause two different clusters to get joined. This observation allows us to have a robust stopping criterion.

We describe our output in Table~\ref{tab:PC-intersect-C},
where the $(i,j)$-th entry indicates $|PC_i \cap V_j |$. 
On a high level one can observe that $V_5$ is broken down into 
two clusters, and the rest of the clusters in $V$ have one corresponding primary cluster. Furthermore, the misclassification error 
is \emph{less than $1\%$} for $6$ out of $7$ primary clusters, 
with one $PC_3$ having $2\%$ error. 
The purity of primary clustering is one of the motivations of this work. We describe the significance of this result in Section~\ref{sec:4}
in more detail.

\begin{table}[H]
    \centering
\begin{tabular}{|c|c|c|c|c|c|c|} \hline
 & $V_1$ & $V_2$ & $V_3$ & $V_4$ & $V_5$ & $V_6$ \\ \hline 
$PC_1$ &  0  &  0  &  0  &  0  &  465  &  0 \\ \hline
$PC_2$  &  4  &  783  &  0  &  0  &  3  &  1 \\ \hline
$PC_3$  &  0  &  0  &  0  &  889  &  14  &  0 \\ \hline
$PC_4$  &  0  &  1  &  0  &  0  &  1  &  650 \\ \hline
$PC_5$  &  917  &  6  &  0  &  0  &  0  &  0 \\ \hline
$PC_6$  &  0  &  0  &  174  &  0  &  0  &  0 \\ \hline 
 \end{tabular}
    \caption{Primary clustering for $\gamma=0.05$ with $867$ remaining vertices}
    \label{tab:PC-intersect-C}
\end{table}

\paragraph{Robustness w.r.t $\delta$}
We observe that the merge procedure is robust w.r.t $\delta$. In fact, we obtain the same clustering for $1 \le \delta \le 5$.

\subsection{Final clustering via majority voting}
Once we have the primary clustering, we cluster the remaining vertices. These are the vertices that did not form any confident set in Algorithm~\ref{alg:confident-set}. In this paper we apply a relatively simple algorithm, that we now describe.

\begin{algorithm}[!htb]
\caption{\textsc{Clustering the remaining vertices}}\label{alg:finish-cluster}
\begin{algorithmic}[1]
\STATE Take an input $S$ and $CS=\{CS_1, \ldots CS_r \}$,
and $\{N_{\bu} \}_{\bu \in P}$ from
Algorithm~\ref{alg:confident-cluster}.
\STATE $t \gets \frac{\delta n}{100}$ 
\vspace{0.1cm}
\STATE $P \gets S\setminus CS_1\cup\cdots\cup CS_r$ \hfill [Currently unclassified vertices]
\FOR{ $1 \le i \le r$}
\STATE $CS'_i \gets \emptyset$
\ENDFOR 
\FOR{$\bu \in P$}
\STATE $SP(\bu) \gets \{j : |N_{\bu} \cap CS_j| > t/2 \}$ \hfill [$SP$ has cardinality at most $1$]
\IF{$SP(\bu) \neq \emptyset$}
\STATE Let $\ell \gets SP(\bu)$ 
\STATE $CS'_{\ell} \gets CS'_{\ell} \cup \{\bu\}$
\ENDIF
\ENDFOR \hfill 
\FOR{ $1 \le i \le r$}
\STATE $CC_i \gets CS'_i \cup CS_i$
\ENDFOR 
\STATE Output $CC$ \hfill [Some vertices may still be unclassified]
\end{algorithmic}
\end{algorithm} 

The method is simple. For every unclustered vertex, we look at its 
$t=\frac{\delta n}{100}$, that is top $\delta \%$ of the neighbours.
If more than half of these neighbours belong to any of the primary cluster, we add it to the cluster, otherwise we move on to the next vertex. In the end we have our second layer of clustering output,
and some leftover vertices (vectors). The rationale behind this step is simple. The vertices that did not become part of any clique did not have strong signals w.r.t to the approximate correlation graph in our algorithm. We add these to a primary cluster if they contain majority of its top neighbours. For dataset-1, after Algorithm~\ref{alg:confident-cluster} has finished, we have $867$ (out of a total $4743$) vectors that are passed on. After this step, depending on the choice of $\delta$, it can become as low as $73$. 
This step increases the misclassification error at the cost of inclusion of vectors that have comparatively weaker signal, but clusters many vertices. 
For this case we have the following choices and outcomes.
\begin{table}[H]
    \begin{center}
    \begin{tabular}{| c | c |} \hline
        Property & Value  \\ \hline
        Number of total vectors & $4743$  \\ \hline
        Choice of $k'$ & 20 \\ \hline
         Choice of $\gamma$ & 5 \\ \hline
         Choice of $\delta$ & $2.5$ \\ \hline
         Number of clusters formed  & $6$ \\ \hline
          Remaining vertices & $69$ \\ \hline
    \end{tabular}
    \end{center}
    \caption{Status after Algorithm~\ref{alg:finish-cluster} for dataset-1}
    \label{tab:my_label}
\end{table}

The remaining few vectors can be either added to the cluster with highest number of neighbours from $N_{\bu}$, or we can run the algorithm recursively on it to identify a small cluster. The choice is left to the user. For our experiment, we take the former course, and describe the outcome in Section~\ref{sec:4}.

%%%%%%%%%%%%%%%%%%%%%%%%%%%

\section{Analysis of output and comparison with Seurat \label{sec:4}}
In Section~\ref{sec:2} and \ref{sec:3} we have described our algorithm with theoretical background, and also demonstrated some step wise outputs for dataset-1.
Here we compare our outputs with the (almost) true  labels provided by Zheng et. al.~\cite{zheng2017massively} and Seurat's output in several ways, underpinning the important characteristics of our algorithm. 
As we have described in Section~\ref{sec:3-dataset}, the dataset consists of $6$ clusters, with their labels obtained from~\cite{zheng2017massively}. The original size of the clusters in the dataset are ($11213, 10209, 2612, 10085, 9232, 8385$), 
and we choose cells from each of these clusters with probability $\frac{1}{10}$. For this particular experiment, we use dataset-1, where the number cells in each of the clusters are $(1022,924,260,955,839,743)$, with correspondence to CD4, CD8, CD14, CD19, CD34 and CD56 respectively. 

We denote the labeling of ~\cite{zheng2017massively} as $V_1, \ldots ,V_k$ (where $k=6$), 
our primary clustering with $PC= \{PC_1, \ldots ,PC_{r_1}\}$, post-majority clustering 
as $C=\{C_1, \ldots ,C_{r_1}\}$ and Seurat's clustering as $SC=\{SC_1, \ldots ,SC_{r_2} \}$ 
where $r_1$ and $r_2$ are the number of clusters formed by our algorithm and Seurat, respectively. 
We concentrate on comparisons on two metrics,
accuracy and robustness w.r.t input parameters, which are two of the most important metrics of analysis of unsupervised learning algorithms.

\subsection{Accuracy \label{sec:4-accuracy}}
We first define our notion of accuracy, which we compare with the 
ground truth clusters $V=\{V_1, \ldots ,V_k\}$ provided by \cite{zheng2017massively}. 
Here we take into account that the number of clusters
output by the algorithms may not be the same as that of the ground truth clusters, as the algorithms can 
break down a ground truth cluster into some sub-clusters, as there may be underlying hierarchy in the dataset. 

\begin{definition}[Error of one cluster]

\item Given a clustering $C=\{ C_1, \ldots , C_r\}$,
we define the primary identity of $C_j$ w.r.t $V$
as 
\[I(C_j)= \argmax_{1 \le i \le k} \left| C_j \cap V_i  \right| \]
Then the accuracy of $C_i$ w.r.t $V$ is defined as 
$e(C_j,V)= \frac{\left| C_j \setminus V_{I(C_j)} \right|}{|C_j|}$. 
Thus, we define the identity of a cluster $C_i$
as the majority ground truth cluster contained in it. And then calculate its misclassification fraction of cells (vectors) in $C_j$ that are not of $V_i$. 
\end{definition}

Then we have the two following forms of error.

\begin{definition}[Error in clustering]
\ 

\begin{enumerate}
    \item We define $e_{\infty}(V,C)= \max_{ 1 \le j \le r} e(C_j,V)$. This is the maximum cluster wise error. 
    
    \item We define $e_{avg}(V,C) =\sum_{1 \le j \le r} e(C_j,V)$. This is the average cluster-wise error. Here note that the error is already weighted according to the cluster size.

\end{enumerate}

\end{definition}

We first note down the step-wise outcomes of our algorithm for
our first setting of parameters, with $\gamma=0.05, \delta=2.5$ and $k'=20$, and then compare it with Seurat.

\begin{table}[H]
    \centering
\begin{tabular}{|c|c|c|c|c|c|c|c|} \hline
 & $V_1$ & $V_2$ & $V_3$ & $V_4$ & $V_5$ & $V_6$ & $e(PC_j,V)$\\ \hline 
$PC_1$ &  0  &  0  &  0  &  0  &  465  &  0 & 0\\ \hline
$PC_2$  &  4  &  783  &  0  &  0  &  3  &  1 & 0.01\\ \hline
$PC_3$  &  0  &  0  &  0  &  889  &  14  &  0 & 0.015\\ \hline
$PC_4$  &  0  &  1  &  0  &  0  &  1  &  650 & 0.003\\ \hline
$PC_5$  &  917  &  6  &  0  &  0  &  0  &  0 & 0.006\\ \hline
$PC_6$  &  0  &  0  &  174  &  0  &  0  &  0 & 0\\ \hline 
 \end{tabular}
    \caption{Primary clustering for $\gamma=0.05$ with $867$ remaining vertices}
    \label{tab:PCgamma05k20}
\end{table}

\begin{table}[H]
    \centering
\begin{tabular}{|c|c|c|c|c|c|c|c|} \hline
 & $V_1$ & $V_2$ & $V_3$ & $V_4$ & $V_5$ & $V_6$ & $e(PC_j,V)$  \\  \hline
$C_1$ &  3  &  1  &  13  &  1  &  724  &  1 & 0.02\\ \hline
$C_2$  &  4  &  879  &  0  &  0  &  5  &  9 & 0.02\\ \hline
$C_3$  &  0  &  0  &  0  &  951  &  33  &  1 & 0.034\\ \hline
$C_4$  &  0  &  6  &  0  &  0  &  2  &  728 & 0.0135\\ \hline
$C_5$  &  1009  &  22  &  2  &  1  &  0  &  0 & 0.024\\ \hline
$C_6$  &  0  &  0  &  237  &  0  &  42  &  0 & 0.15\\ \hline
 \end{tabular}
    \caption{Post-majority clustering for $\gamma=0.05$  and 69  remaining vertices}
    \label{Cgama05k20}
\end{table}

Now we compare our result with several Seurat outcomes. 
Seurat needs two inputs, PCA projection dimension and resolution. 
We choose number of PCs as $20$, where this choice is heuristic and
Seurat suggests a number between $10$ to $30$. 
We choose the resolution based on the following remark of Seurat tutorial.
\begin{quote}
    \emph{``We find that setting this parameter between 0.4-1.2 typically returns good results for single-cell datasets of around 3K cells. 
    Optimal resolution often increases for larger datasets.''}
\end{quote}
Additionally Seurat outputs a modularity value, where a higher modularity
implies the graph clustering captures more of the structure of the graph. 
We found that in the range of $0.4$-$1.2$, the modularity is highest for $0.4$. 

\begin{table}[H]
    \centering
\begin{tabular}{|c|c|c|c|c|c|c|c|} \hline
 & $V_1$ & $V_2$ & $V_3$ & $V_4$ & $V_5$ & $V_6$ & $e(S'_j,V)$ \\ \hline 

$SC_1$ &  1013  &  6  &  2  &  1  &  0  &  1 & 0.00978 \\ \hline
$SC_2$  &  0  &  0  &  3  &  954  &  \cred{30}  &  0 & 0.0334 \\ \hline
$SC_3$  &  3  &  3  &  2  &  0  &  3  &  732 &  0.0148 \\ \hline
$SC_4$  &  0  &  485  &  0  &  0  &  4  &  1 & 0.0102\\ \hline
$SC_5$  &  4  &  430  &  0  &  0  &  0  &  9 & 0.0293\\ \hline
$SC_6$  &  0  &  0  &  0  &  0  &  410  &  0 & 0\\ \hline
$SC_7$  &  0  &  0  &  236  &  0  &  \cred{100}  &  0 & \cred{0.297}\\ \hline
$SC_8$  &  2  &  0  &  \cred{17}  &  0  &  259  &  0 & 0.0683\\ \hline
$SC_9$  &  0  &  0  &  0  &  0  &  33  &  0 & 0\\ \hline
\end{tabular}
    \caption{Seurat clustering for $20$-PC and resolution=$0.4$ with modularity=$0.9485$}
    \label{tab:SC09}
\end{table}

Now, we note the following observations, that highlight the consistency of primary clustering, which we argue as the one of the main motivations behind this algorithm.

\begin{observation}
The cluster $V_4$ has $955$ cells.
Its majority cluster in Seurat clustering is $SC_2$, that has 
$954$ cells from $V_4$ and $30$ cells from $V_5$.

Its majority cluster in primary clustering is $PC_2$,
$889$ cells from $V_4$ and $14$ cells from $V_5$.
In addition, we have 
\[  
PC_2 \setminus V_4 \subseteq SC_2\setminus V_4
\]
It implies the $14$ cells in $PC_2$ disagreeing with $V_4$
is a subset of cells in $SC_2$ disagreeing with $V$. 
\end{observation}

Next we look at $SC_7$. This cluster is the major erroneous cluster in Seurat w.r.t $V$, as it has a large mixture of 
$V_3$ and $V_5$ cells. In this direction we have the following
observation.

\begin{observation}
$V_3$ is the smallest cluster in our dataset with $260$ cells.
Its majority cluster in Seurat is $SC_7$, that $232$ many $V_3$ cells  but also $106$ many $V_5$ cells. 

In primary clustering, the majority cluster of $V_3$ is $PC_6$, with $176$ many $V_3$ cells and no miscalssification. Here we also have
\[ 
PC_6 \subset SC_7 \cap V_3
\]
This implies that the $174$ out of $260$ cells of $V_3$
clustered in $PC_6$ is a subset of the $232$ cells of $SC_7$, and \bf{our primary clustering is consistent with both  \cite{zheng2017massively}'labels and Seurat.}

\end{observation}

For this setting, we found that in $SC_7$, $V_3$ and $V_5$ are highly mixed, 
giving an error of $0.29762$, or $29\%$, which is much worse than even
our post majority clustering. Now, we increase $res=0.5$, which gives us $11$
clusters (where there are $6$ true clusters). This breaks down $SC_7$ into two multiple clustersm reducing the mixing. Let us observe the output.

Thus, we observe Seurat's output for resolution=$0.5$ and $20$ PCs. 
 \begin{table}[H]
    \centering
\begin{tabular}{|c|c|c|c|c|c|c|c|} \hline
 & $V_1$ & $V_2$ & $V_3$ & $V_4$ & $V_5$ & $V_6$ & $e(SC'_j,V)$ \\ \hline 

$SC'_1$ &  1013  &  6  &  2  &  1  &  0  &  0 & 0.0185  \\ \hline
$SC'_2$ &  3  &  3  &  2  &  0  &  3  &  732 & 0.0148   \\ \hline
$SC'_3$ &  0  &  0  &  0  &  711  &  21  &  0 & 0.0286  \\ \hline
$SC'_4$ &  0  &  485  &  0  &  0  &  4  &  1 & 0.01     \\ \hline
$SC'_5$ &  4  &  430  &  0  &  0  &  0  &  10 & 0.031   \\ \hline
$SC'_6$ &  0  &  0  &  0  &  0  &  410  &  0 & 0.047    \\ \hline
$SC'_7$ &  0  &  0  &  3  &  243  &  9  &  0 & 0.052    \\ \hline
$SC'_8$ &  2  &  0  &  10  &  0  &  232  &  0 & 0.049   \\ \hline
$SC'_9$ &  0  &  0  &  230  &  0  &  1  &  0 &   0.004   \\ \hline
$SC'_{10}$ &  0  &  0  &  13  &  0  &  126  &  0   & \cred{0.0928} \\ \hline
$SC'_{11}$ &  0  &  0  &  0  &  0  &  33  &  0 & 0      \\ \hline
\end{tabular}
    \caption{Seurat clustering for $20$-PC and resolution=$0.5$ with modularity=$0.93$}
    \label{tab:SC11-20PC-res05}
\end{table}

Here, at the expense of a higher number of clusters, Seurat's error decreases. However, $e_{\infty}$ is still higher compared to primary clustering,
as there is still comparatively high mixing between cluster $3$ and cluster $5$. The $e_{avg}$ error of our post majority clustering also remains better then Seurat for both the $9$ cluster and $11$ cluster outputs. This is because Seurat has more, small clusters, so the fractional error values are higher.

\begin{table}[H]
    \centering
    \begin{tabular}{|c|c|c|c|c|c|} \hline 
        Clustering & Number of clusters &  Parameters &$e_{\infty}$ & $e_{avg}$ & Unclustered points\\ \hline 
         CC & 7 & $\gamma$=0.05 &  \color{teal}{0.0155} &  \color{teal}{0.034} & 867\\ \hline 
         C & 7 & $\gamma$=0.05 & 0.15 &  0.2615 & 69\\ \hline 
         SC & 9 & res=0.4 & 0.293 &  0.463 & 0\\ \hline 
         SC' & 11 & res=0.5 & 0.0928 &  0.3477 & 0\\ \hline
     \end{tabular}
    \caption{Comparing the accuracy of our algorithm and Seurat for resolution $0.4$ and $0.5$ }
    \label{tab:seurat-20-05}
\end{table}

\paragraph{Takeaway}
\begin{enumerate}
    \item The main takeaway is that the misclassification of primary clustering remains really low. We obtain this in exchange of having a high number of unclustered points. If a cell is clustered during the primary clustering phase, then it guarantees that it is part of a densely connected cluster w.r.t an overall PCA compression, and this allows us to have really low $e_{\infty}$ and $e_{avg}$ errors. 

    \item Seurat mixes $V_3$ and $V_5$ for both the resolution choices. The corresponding cell types are
    monocytes (CD14) and hematopoietic stem cells  (CD34). Hematopoietic stem cells are a common ancestor to other kind of cells, which makes it plausible that there are intermediate cells in this dataset that are in between monocytes and hematopoietic stem cells. 

    Against this backdrop, our confident clustering approach finds a subset of cells which it can clearly separate, and once this separation is complete, it also promises to help cluster the ``intermediate'' cells more accurately. It should be interesting to observe the performance of our algorithm on datasets with different kind of boundary/ intermediate cells.

    \item After our majority voting phase, we cluster all the vertices, which also increases the misclassification error. Still the errors remain 
    comparable to Seurat's results, which is a reflection of the previous point. 

\end{enumerate}

\subsection{Robustness \label{sec:4-robust}}
Having described the accuracy  and consistency of our algorithm, now we describe its robustness. Unsupervised algorithms 
generally have some heuristic/user input parameter that are chosen 
based on prior knowledge of the problem at hand. For example,
Seurat has two parameters that are chosen by the user, which is the number of PCs for projection and resolution. 
Additionally, when it creates the nearest neighbour graph, it looks at $20$-neighborhood. 

Similarly, we have two variable parameters, $k'$(same as Seurat) 
and $\gamma$, where we build our approximation correlation graph
based on the top $\gamma \%$ of the highest compressed edges.
Additionally, we choose $\delta$, which determines the neighborhood 
for the merging step. In the next part, we always choose $\delta=2.5 \%$.

\subsubsection{Sensitivity w.r.t $\gamma$ and $k'$}

Now we show that our primary clustering is robust to a range of choices of $k'$ and $\gamma$. First we make the following observation about $\gamma$.

\begin{remark}
When we choose $\gamma$, a higher value indicates higher $\alpha$ and $\beta$ value in the $(\alpha,\beta)$-corresponding graph, whereas 
a lower value of $\gamma$ decreases both $\alpha$ and $\beta$.
\end{remark}
In this direction, choosing high $\gamma$ is not advisable, as it increase $\beta$, which is the percentage of ``inter-cluser'' edges, 
then we risk having errors in the confident sets. 

On the other hand, low $\gamma$ may mean there are not sufficiently enough edges for there to be meaningful dense subgraphs. Thus, when $\gamma$ is low, we expect a higher number of unclustered cells. Here we look at two $\gamma$ value, $0.03$ and $0.05$. 

Next, we discuss the dimension of the PCA projection. As we have discussed above, \cite{mukherjee2022compressibility} showed that  the best choice of dimension for relative PCA compression is the number of underlying clusters, they further showed that the result is not very sensitive to the choice of PCs.

However, the choice of PC can still effect the outcome of algorithms that uses PCA projection as a tool, such as Seurat. For example, if we choose $PC=10$ and $20$ for the same choice of resolution=$0.5$, (which is a standard input for resolution), we get different outcomes. 
In fact, we have a very interesting observation.

\begin{observation}
Seurat's error of mixing w.r.t to the the PCA projection dimension is not monotonic. For $k'=20$, it generates $11$ clusters, with $e_{\infty}$ error of around $0.09$, but for both $k'=10$ and $k'=30$
it has $9$ and $10$ clusters respectively, both having $>0.25$ maximum error. 
\end{observation}
We describe these results in Tables~\ref{tab:seurat-20-05}, \ref{tab-tab-seurat-10,05} and \ref{tab-seurat-30,05}. 
We summarize this in the following Table~\ref{tab:seurat-diff-k}. 
\begin{table}[H]
    \centering
    \begin{tabular}{|c|c|c|c|c|} \hline 
         $res$ & $k'$ &$e_{\infty}$ & $e_{avg}$ & Number of clusters \\ \hline 
          0.5 & 10 & 0.308   &   0.56  &  9 \\ \hline 
          0.5 & 20 & 0.0928 &   0.3477   &  11 \\ \hline
          0.5 & 30 & 0.338   &   0.967 & 10 \\ \hline
     \end{tabular}
    \caption{Output of Seurat for same resolution and different $k'$}
    \label{tab:seurat-diff-k}
\end{table}

\begin{table}[H]
    \centering
\begin{tabular}{|c|c|c|c|c|c|c|c|} \hline
 & $V_1$ & $V_2$ & $V_3$ & $V_4$ & $V_5$ & $V_6$ & $e(PC_j,V)$ \\ \hline 
$SC_1$ &  1013  &  10  &  2  &  0  &  0  &  0 & 0.012\\ \hline
$SC_2$ &  0  &  0  &  0  &  728  &  22  &  0 & 0.029\\ \hline
$SC_3$ &  0  &  4  &  0  &  0  &  2  &  717 & 0.008\\ \hline
$SC_4$ &  1  &  503  &  0  &  0  &  3  &  1 & 0.009\\ \hline
$SC_5$ &  3  &  1  &  1  &  0  &  439  &  5 & 0.022\\ \hline
$SC_6$ &  3  &  406  &  1  &  1  &  0  &  20 & 0.058\\ \hline
$SC_7$ &  0  &  0  &  236  &  0  &  105  &  0 & 0.308\\ \hline
$SC_8$ &  2  &  0  &  17  &  0  &  260  &  0 & 0.068\\ \hline
$SC_9$ &  0  &  0  &  3  &  226  &  8  &  0 & 0.046\\ \hline
 \end{tabular}
    \caption{Seurat Clustering for $k'=10$, $res=0.5$, creating $9$ clusters}
    \label{tab-tab-seurat-10,05}
\end{table}

\begin{table}[H]
    \centering
\begin{tabular}{|c|c|c|c|c|c|c|c|} \hline
 & $V_1$ & $V_2$ & $V_3$ & $V_4$ & $V_5$ & $V_6$ & $e(PC_j,V)$ \\ \hline 
$SC_1$ &  1011  &  5  &  2  &  1  &  0  &  1 & 0.009\\ \hline
$SC_2$ &  2  &  4  &  2  &  0  &  3  &  731 & 0.012\\ \hline
$SC_3$ &  0  &  0  &  0  &  690  &  21  &  0 & 0.029\\ \hline
$SC_4$ &  0  &  491  &  0  &  0  &  3  &  1 & 0.008\\ \hline
$SC_5$ &  7  &  424  &  0  &  0  &  0  &  10 & 0.038\\ \hline
$SC_6$ &  0  &  0  &  0  &  0  &  412  &  0 & 0\\ \hline
$SC_7$ &  0  &  0  &  243  &  0  &  124  &  0 & 0.338\\ \hline
$SC_8$ &  0  &  0  &  3  &  264  &  9  &  0 & 0.043\\ \hline
$SC_9$ &  2  &  0  &  10  &  0  &  232  &  0 & 0.49\\ \hline
$SC_{10}$ &  0  &  0  &  0  &  0  &  35  &  0 & 0\\ \hline
 \end{tabular}
    \caption{Seurat Clustering for $k'=30$, $res=0.5$, creating $10$ clusters}
    \label{tab-seurat-30,05}
\end{table}

Now we describe the outcome of our algorithm for $k'=10,20$ and $30$.
We show that our output is fairly stable across these choices of $\gamma$ and $k'$(PCA projection dimension), verifying our algorithm's robustness. We document the summary in Table~\ref{tab:robust-gamma} and the details in Tables~\ref{tab-robust-1},\ref{tab-robust-2},\ref{tab-robust-3} and \ref{tab-robust-4}.

\begin{table}[H]
    \centering
    \begin{tabular}{|c|c|c|c|c|c|} \hline 
         Number of clusters & $\gamma$ & $k'$ &$e_{\infty}$ & $e_{avg}$ & Unclustered points\\ \hline 
         
         7 & 0.03 & 20 & 0.004 & 0.007 &  1244 \\ \hline 
         7 & 0.03 & 10 & 0.007 & 0.025  &  1005 \\ \hline
         7 & 0.05 & 10 & 0.019 & 0.062 & 793 \\ \hline
         7 & 0.05 & 30 & 0.018 & 0.031 & 954 \\ \hline
     \end{tabular}
    \caption{Stability of primary clustering w.r.t $\gamma$ and $k'$ }
    \label{tab:robust-gamma}
\end{table}
Here we show, that even for several different combinations of $k'$ as well as $\gamma$, the total cells (vectors) we select into confident sets are of comparable size, with the 
primary clustering error also being similar, and low. 
We document the results in the following table.

\begin{table}[h]
    \centering
    \begin{center}
    \begin{tabular}{|c|c|c|c|c|c|c|c|} \hline
 & $V_1$ & $V_2$ & $V_3$ & $V_4$ & $V_5$ & $V_6$ & $e(PC_j,V)$ \\ \hline 
$PC_1$  &  1  &  704  &  0  &  0  &  0  &  0 & 0.001\\ \hline
$PC_2$ &  0  &  0  &  0  &  803  &  3  &  0 & 0.004\\ \hline
$PC_3$ &  0  &  0  &  0  &  0  &  295  &  0 & 0\\ \hline
$PC_4$ &  0  &  0  &  0  &  0  &  1  &  570 & 0.002\\ \hline
$PC_5$ &  843  &  0  &  0  &  0  &  0  &  0 & 0\\ \hline
$PC_6$ &  0  &  0  &  170  &  0  &  0  &  0 & 0\\ \hline
$PC_7$ &  0  &  0  &  0  &  0  &  129  &  0 & 0\\ \hline
     \end{tabular}
        \end{center}
    \caption{Primary clusters: $\gamma=0.03$, $k'=20$ with {\bf 1244} many unassigned cells}
    \label{tab-robust-1}
\end{table}

\begin{table}[H]
    \centering
\begin{tabular}{|c|c|c|c|c|c|c|c|} \hline
 & $V_1$ & $V_2$ & $V_3$ & $V_4$ & $V_5$ & $V_6$ & $e(PC_j,V)$ \\ \hline 
 $PC_1$ &  1  &  767  &  0  &  0  &  0  &  8 & 0.012\\ \hline
 $PC_2$  &  0  &  0  &  0  &  836  &  6  &  0 & 0.007\\ \hline
 $PC_3$  &  888  &  1  &  0  &  0  &  0  &  0 & 0.001\\ \hline
 $PC_4$  &  0  &  0  &  0  &  0  &  320  &  0 & 0\\ \hline
 $PC_5$  &  0  &  1  &  0  &  0  &  2  &  625 & 0.005\\ \hline
 $PC_6$  &  0  &  0  &  172  &  0  &  0  &  0 & 0\\ \hline
 $PC_7$  &  0  &  0  &  0  &  0  &  131  &  0 & 0\\ \hline
\end{tabular}
    \caption{Primary clusters: $\gamma=0.03$, $k'=10$ with {\bf 1005} many unassigned cells}
    \label{tab-robust-2}
\end{table}

\begin{table}[H]
    \centering
\begin{tabular}{|c|c|c|c|c|c|c|c|} \hline
 & $V_1$ & $V_2$ & $V_3$ & $V_4$ & $V_5$ & $V_6$ & $e(PC_j,V)$ \\ \hline 
$PC_1$ &  8  &  810  &  0  &  0  &  3  &  12 &  0.018  \\ \hline
$PC_2$ &  0  &  0  &  0  &  0  &  340  &  0 &   0  \\ \hline
$PC_3$ &  0  &  0  &  0  &  910  &  18  &  0 &  0.019  \\ \hline
$PC_4$ &  0  &  12  &  0  &  0  &  1  &  644 &  0.019  \\ \hline
$PC_5$ &  930  &  5  &  1  &  0  &  0  &  0 &   0.006  \\ \hline
$PC_6$ &  0  &  0  &  176  &  0  &  0  &  0 &   0  \\ \hline
$PC_7$ &  0  &  0  &  0  &  0  &  111  &  0 &   0  \\ \hline
\end{tabular}
    \caption{Primary clusters: $\gamma=0.05$, $k'=10$ with {\bf 793} many unassigned cells}
    \label{tab-robust-3}
\end{table}

\begin{table}[H]
    \centering
\begin{tabular}{|c|c|c|c|c|c|c|c|} \hline
 & $V_1$ & $V_2$ & $V_3$ & $V_4$ & $V_5$ & $V_6$ & $e(PC_j,V)$ \\ \hline 
$PC_1$ &  0  &  0  &  0  &  878  &  16  &  0 & 0.018\\ \hline
$PC_2$  &  2  &  765  &  0  &  0  &  2  &  0  & 0.005\\ \hline
$PC_3$  &  0  &  0  &  0  &  0  &  309  &  0  & 0\\ \hline
$PC_4$  &  0  &  1  &  0  &  0  &  2  &  613  & 0.005\\ \hline
$PC_5$  &  902  &  3  &  0  &  0  &  0  &  0  & 0.003\\ \hline
$PC_6$  &  0  &  0  &  0  &  0  &  152  &  0  & 0\\ \hline
$PC_7$  &  0  &  0  &  176  &  0  &  0  &  0  & 0\\ \hline
 
 \end{tabular}
    \caption{Primary clusters: $\gamma=0.05$, $k'=30$ with {\bf 954} many unassigned cells}
    \label{tab-robust-4}
\end{table}

\subsubsection{Sensitivity w.r.t $\delta$}

Finally, we describe the performance of the merging step w.r.t to choice of $\delta$. As we have described, the user can stop the merging process depending on the value of $Z_{max}$ at that round. Here we note some outcomes, that underlines the stability of choice of $\delta$ in terms of the similarity in the final output if number of primary clusters are same. 

\begin{enumerate}
    \item For $k'=10, \gamma=0.03$, the primary clusters are identical for $\delta \in \{0.01, 0.015, 0.025\}$ when merged until there are $7$ clusters remaining. Furthermore, the $Z_{max}$ for the final and penultimate 
    rounds are similar, whereas it becomes much less for the next round, where we stop. 

    \item For $k'=30$, $\gamma=0.05$, the primary clusters are again identical for  $\delta \in \{0.01, 0.015, 0.025\}$ when merged until there are $7$ clusters remaining.
    
\end{enumerate}

\section{Conclusion and future directions}
\label{sec:5}
In this paper we have designed a novel clustering algorithm with the motivation of confident clustering. Our algorithm is designed using the relative compressibility of PCA, along with spectral techniques that we draw form the dense graph regime. We apply our algorithm on a well known complex dataset~\cite{zheng2017massively}, and obtain a primary clustering result that has very low misclassification error, w.r.t to the label provided by \cite{zheng2017massively}, and importantly, is consistent with the  label from~\cite{zheng2017massively} and the outcome of Seurat, one of the most popular clustering algorithms in this domain. 

We further observe that our algorithm is robust to the choice of number of PCs as compared to Seurat, and is also reasonably stable w.r.t to the other user input parameters. In addition, compared with traditional clustering algorithms, our algorithm is more robust for different parameters.

\subsection{Future directions}

The area of clustering of single-cell sequencing data has many challenging problems, because of the extremely complicated nature of biological data. 
In this paper we have designed an algorithm with the viewpoint of accurate and robust clustering. Some of the other important directions areas are as follows. In this direction, we have several problems that we believe are important in clustering of single-cell data as well as understanding of simple spectral algorithms. 

\begin{enumerate}
    \item {\bf Runtime:} With the improvement in the single-cell sequencing analysis, both the number of cells and the number of genes per cell are increasing. Today there are many datasets with millions of cells, of dimension $50,000$, or more. In such a setting, we can only suffer linear time  (on the size of $\hA$) algorithms. Furthermore, the prospect of online clustering of new cells promises to be a challenging and rewarding direction.
    
    \item {\bf Hierarchical clustering} As we have seen on a high level, the cells present in single-cell data have internal hierarchies. In this direction, modelling single-cell data in a hierarchical manner is an exciting problem, one that should also help with interpretability of clustering result.

\end{enumerate}

Beyond, single-cell analysis, we feel the concept of boundary points in clustering problem deserves attention. Furthermore, analyzing our spectral steps, such as obtaining dense subgraphs via projection onto top eigenvectors in a tight theoretical model promises to further establish the power of simple spectral algorithms. In our algorithm, there are certain constants for which we make a reasonable choice (and one that works for several different choice of other parameters), such as the condition in stopping criterion for confident sets. A more robust understanding of such constants may help us further improving the algorithm w.r.t both the accuracy and robustness metrics.

As an ending note, we plan to collaborate with biologists on analyzing single-cell sequencing data through our clustering algorithms. We hope our algorithm is able to make contributions in this area, which will certainly be the most desirable outcome of this work.

%\bibliographystyle{alpha}
%\bibliography{reference}

\bibliographystyle{alpha}

\end{document}